\pdfoutput=1

\documentclass[11pt]{article}

\usepackage[]{acl}
\usepackage{times}
\usepackage{latexsym}

\usepackage[T1]{fontenc}

\usepackage[utf8]{inputenc}

\usepackage{microtype}

\usepackage{multirow}
\usepackage{booktabs}
\usepackage{hyperref}
\usepackage{amsmath}
\usepackage{framed}  %
\usepackage{array}

\usepackage{graphicx}  %

\newcommand{\ours}{\textsc{NatCS}}
\newcommand{\oursself}{\textsc{NatCS\textsubscript{Self}}}
\newcommand{\oursspoken}{\textsc{NatCS\textsubscript{Spoke}}}
\newcommand{\real}{\textsc{Real}}

\newcommand{\oursbanking}{Banking}
\newcommand{\oursfinance}{Finance}
\newcommand{\ourshealth}{Health}
\newcommand{\ourstravel}{Travel}
\newcommand{\oursinsuranceself}{Insurance}

\newcommand{\realretaila}{Retail\textsubscript{A}}
\newcommand{\realretailb}{Retail\textsubscript{B}}
\newcommand{\realretailc}{Retail\textsubscript{C}}
\newcommand{\realfinancea}{Finance\textsubscript{A}}
\newcommand{\realfinanceb}{Finance\textsubscript{B}}

\newcommand{\sgd}{SGD}
\newcommand{\multidogo}{\textsc{MDGO}}

\newcommand{\multiwoz}{MWOZ}
\newcommand{\tmself}{TM1\textsubscript{Self}}

\newcommand{\semdiv}{SemDiv\textsubscript{intent}}
\newcommand{\ittr}{TTR\textsubscript{intent}}

\newcommand{\ourdas}{\ours~dialogue~acts}
\newcommand{\Ourdas}{\ours~Dialogue~Acts}
\newcommand{\da}[1]{\textit{#1}}
\newcommand{\plusminus}[1]{$ {\scriptscriptstyle\pm{#1}}$}
\newcommand{\std}[2]{#1\plusminus{#2}}

\newcommand\labelbox[2]{\underline{#1}$_{\text{\tiny\textsc{#2}}}$}
\newcommand\ann[1]{#1}
\newcommand{\realood}{Real\textsubscript{OOD}}
\newcommand{\realid}{Real\textsubscript{ID}}

\title{\ours{}: Eliciting Natural Customer Support Dialogues}

\author{
    \textbf{\large James Gung, Emily Moeng, Wesley Rose}\\
    \textbf{\large Arshit Gupta, Yi Zhang, and Saab Mansour}\\
    AWS AI Labs \\
    \texttt{\{gungj,emimoeng,rosewes,arshig,yizhngn,saabm\}@amazon.com}\\
}

\begin{document}
\maketitle
\begin{abstract}
Despite growing interest in applications based on natural customer support conversations, there exist remarkably few publicly available datasets that reflect the expected characteristics of conversations in these settings.
Existing task-oriented dialogue datasets, which were collected to benchmark dialogue systems mainly in written human-to-bot settings, are not representative of real customer support conversations and do not provide realistic benchmarks for systems that are applied to natural data.
To address this gap, we introduce \ours{}, a multi-domain collection of spoken customer service conversations.
We describe our process for collecting synthetic conversations between customers and agents based on natural language phenomena observed in real conversations.
Compared to previous dialogue datasets, the conversations collected with our approach are more representative of real human-to-human conversations along multiple metrics.
Finally, we demonstrate potential uses of \ours{}, including dialogue act classification and intent induction from conversations as potential applications, showing that dialogue act annotations in \ours{} provide more effective training data for modeling real conversations compared to existing synthetic written datasets. 
We publicly release \ours{} to facilitate research in natural dialog systems\footnote{\url{https://github.com/amazon-science/dstc11-track2-intent-induction}.}. %

\end{abstract}

\section{Introduction}

Applications that are applied to human-to-human customer support conversations have become increasingly popular in recent years.
For example, assistive tools that aim to support human agents, provide analytics, and automate mundane tasks have become ubiquitous in industry applications~\cite{AmazonContactLens,GoogleContactCenterAI,MicrosoftContactCenter}.
Despite this growing interest in data processing for natural customer service conversations, to the best of our knowledge, there exist no public datasets to facilitate open research in this area.

Existing dialogue datasets focusing on development and evaluation of task-oriented dialogue (TOD) systems contain conversations that are representative of human-to-bot (H2B) conversations adhering to restricted domains and schemas~\cite{budzianowski-etal-2018-multiwoz,rastogi2020towards}.
Realistic live conversations are difficult to simulate due to the training required to convincingly play the role of an expert customer support agent in non-trivial domains~\cite{chen-etal-2021-action}.
Existing datasets are also primarily written rather than spoken conversations as this modality is cheaper to simultaneously collect and annotate asynchronously through crowdsourcing platforms.

To address these gaps, we present \ours{}, a multi-domain dataset containing English conversations that simulate natural, human-to-human (H2H), two-party customer service interactions.
First, we describe a self-collection dataset, \oursself{}, where we use a strict set of instructions asking participants to write both sides of a conversation as if it had been spoken.
Second, we present a spoken dataset, \oursspoken{}, in which pairs of participants were each given detailed instructions and asked to carry out and record conversations which were subsequently transcribed.
We observe that the resulting conversations in \ours{} share more characteristics with real customer service conversations than pre-existing dialogue datasets in terms of diversity and modeling difficulty.

We annotate a subset of the conversations in two ways:
(1) we collect task-oriented dialogue act annotations, which label utterances that are important for moving the customer's goal forward and
(2) we categorize and label customer goals and goal-related information with an open intent and slot schema, mimicking the process for building a TOD system based on natural conversations.
We find that classifiers trained with the resulting dialogue act annotations have improvements in accuracy on real data as compared to models trained with pre-existing TOD data.

Our main contributions are threefold:
\begin{itemize}
    \item We present \ours{}, a multi-domain dialogue dataset containing conversations that mimic spoken H2H customer service interactions, addressing a major gap in existing datasets.
    \item We show that \ours{} is more representative of conversations from real customer support centers than pre-existing dialogue datasets by measuring multiple characteristics related to realism, diversity and modeling difficulty.
    \item We provide TOD dialogue act and intent/slot annotations on a subset of the conversations to facilitate evaluation and development of systems that aim to learn from real conversations (such as intent and slot induction), empirically demonstrating the efficacy of the dialogue annotations on real data.
\end{itemize}
Our paper is structured as follows: In Section~\ref{Sec:related_work}, we review other dataset collection methods and approaches for evaluating dialogue quality.
In Section~\ref{Sec:methods}, we describe how \ours{} conversations and annotations were collected.
In Section~\ref{Sec:analysis}, we compare \ours{} with pre-existing dialogue datasets as well as real H2H customer service conversations.
Finally, in Section~\ref{Sec:applications}, we further motivate the dataset through two potential downstream applications, task-oriented dialogue act classification and intent induction evaluation.

\begin{table*}[tbh]
\centering
\addtolength{\tabcolsep}{-3pt}
\scalebox{0.85}{
\begin{tabular}
{p{0.7\linewidth}>{\centering\arraybackslash}m{0.3\linewidth}}
\toprule
 \centering \textbf{Text} & \textbf{Annotations} \\
\midrule
 \textbf{A}: Thank you for calling Intellibank. What can I do for you today? & \ann{ElicitIntent} \\
 \textbf{C}: Hi, I was trying to figure out how to create an account online, but I'm having some trouble. & \ann{InformIntent} \ann{(SetUpOnlineBanking)} \\
 \textbf{A}: I'm sorry to hear that. I'd be happy to help you create an online account. & \\
 \textbf{C}: Well, actually, the reason I called is to check the balance on my account. Could you help me with that? & \ann{InformIntent} \ann{(CheckAccountBalance)} \\
\textbf{A}: Oh, I see. Sure. Let me pull up your account so I can check your balance. May I have your account \labelbox{number}{AccountNumber}, please? & \ann{ElicitSlot} \\
 \textbf{C}: Sure. It's \labelbox{two one two three four five}{AccountNumber}. & \ann{InformSlot} \\
 \textbf{A}: Thanks. And is this account a \labelbox{checking}{TypeOfAccount} or \labelbox{savings}{TypeOfAccount}? & \ann{ConfirmSlot}, \ann{ElicitSlot} \\
 \textbf{C}: Oh, it's \labelbox{checking}{TypeOfAccount}. & \ann{InformSlot} \\
\bottomrule
\end{tabular}
}
\caption{Example conversation from \ours{} \oursbanking{}. Conversations are annotated with dialogue act annotations such as \ann{InformIntent} and \ann{ElicitIntent}, intents such as \ann{SetUpOnlineBanking}, and slots such as \ann{AccountNumber}.}
\label{tab:example_conversation_i}
\end{table*}
\section{Related Work}
\label{Sec:related_work}

\paragraph{Dialogue Dataset Collection}
The goal of \ours{}, to produce conversations that emulate real spoken customer service interactions, differs substantially from previous synthetic dialogue dataset collections.
Previous synthetic, goal-oriented datasets have used the \textit{Wizard of Oz} framework \cite{kelley1984iterative}.
This framework calls for one person to interact with what they think is a computer, but is actually controlled by another person, thus encouraging a human-to-bot (H2b) style.
\citet{wen2016network} define a specific version of this approach to be used with crowdsourced workers to produce synthetic, task-oriented datasets.
This approach has since been adopted as a standard method to collect goal-oriented conversations~\cite{peskov-etal-2019-multi,byrne-etal-2019-taskmaster,budzianowski-etal-2018-multiwoz,el-asri-etal-2017-frames,eric-etal-2017-key}. 

MultiDoGO \cite{peskov-etal-2019-multi} (\multidogo{}), \sgd{} \cite{rastogi2020towards}, and TaskMaster \cite{byrne-etal-2019-taskmaster} are particularly relevant comparisons to our work.
\multidogo{} explicitly encourages dialogue complexities, which serve as inspiration for our complexity-driven methodologies.
\sgd{} is interesting partially because of the scale of their collection, relevance of their task-oriented dialogue act and intent/slot annotations, and because they diverge from the common practice of using the Wizard-of-Oz methodology through the use of dialogue templates and paraphrasing.
TaskMaster presents a methodology for self-collected dialogues, which is analogous to our \oursself{} collection.

Despite these similarities, the methodology for \ours{} differs significantly from any of these pre-existing datasets, because of both the target modality (spoken, or spoken-like conversations), and the setting (H2H instead of H2B).

\paragraph{Analysis of Dialogue Quality}
An important component of our work is the comparison of synthetic dialogue datasets with real data.
We adopt multiple metrics for comparing \ours{} with both real and previous TOD datasets.
\citet{byrne-etal-2019-taskmaster} use perplexity and BLEU score as stand-ins for “naturalness”, with the logic that a dataset should be harder for a model to learn if it is more realistic, because realistic data tends to be more diverse than synthetic data.
Previous collections of dialogue datasets also have made comparisons based on surface statistics such as the number of dialogues, turns, and unique tokens.
\citet{casanueva-etal-2022-nlu} compares intent classification datasets based on both lexical diversity metrics and a semantic diversity metric computed using a sentence embedding model.

There are a number of metrics more common in other sub-domains that may be useful for measuring naturalness in dialogues.
The measure for textual lexical diversity (MTLD), discussed in depth by \citet{mccarthy2010mtld}, provides a lexical diversity measure less biased by document length.
\citet{liao2017measure} introduce a dialog complexity metric, intended for analyzing real customer service conversations, which is computed by assigning importance measures to different terms, computing utterance-level complexity, and weighting the contribution of utterances based on their dialogue act tags.
\citet{hewitt-beaver-2020-case} performed a thorough comparison of the style of human-to-human vs. human-to-bot conversations, using lexical diversity measures, syntactic complexity, and other dimensions like gratitude, sentiment and amount of profanity, though the data was not released.

\section{Methodology}
\label{Sec:methods}

\subsection{Collection Methods}
We propose two collection methods as part of \ours{}. We have three goals for the resulting conversations: (1) They should exhibit the spoken modality, (2) all conversations from each domain should seem to be from the same company, and (3) they should appear to be real, human-to-human conversations between a customer and an agent. We explore two methodologies, resulting in the \oursspoken{} and \oursself{} datasets, to weigh collection cost and complexity compared to dataset effectiveness.

To support goal (3), we propose a set of \textbf{discourse complexity} types.
The motivation for providing specific discourse complexities is to encourage some of the noise and non-linearity present in real human-to-human conversations.
Based on manual inspection of 10 transcribed conversations from a single commercial call center dataset, we identify a combination of human expressions (social niceties, emotionally-charged utterances), phenomena mimicking imperfect, non-linear thought processes (change of mind, forgetfulness/unknown terminology, unplanned conversational flows), reflections of the wider context surrounding the conversation  (continuing from a previous conversation, pausing to find information), distinctions between speakers’ knowledge bases, and the use of multiple requests in single utterances (stating multiple intents, providing multiple slot values).
A list of these complexities along with estimated target percentages (minimum percent of conversations where these phenomena should be present) is provided in Figure~\ref{Fig:discourse}. Descriptions and examples are provided in Appendix Table~\ref{Tab:discourseexamples}.

\begin{figure}[tbh]
\begin{framed}
ChitChat (60\%),
FollowUpQuestion (30\%),
ImplicitDescriptiveIntent (30\%),
PauseForLookup (30\%),
BackgroundDetail (25\%),
MultiElicit (25\%),
SlotCorrection (20\%),
Overfill (15\%),
IntentChange (10\%),
MultiIntent (10\%),
MultiIntentUpfront (10\%),
MultiValue (10\%),
SlotChange (10\%),
Callback (5\%),
Frustration (5\%),
MissingInfoCantFulfill (5\%),
MissingInfoWorkaround (5\%),
SlotLookup (5\%)
\end{framed}
\caption{Discourse complexities and target percentages of conversations containing them used to encourage phenomena observed in real conversations. Percentages were estimated based on manual inspection of a small set of real conversations.}
\label{Fig:discourse}
\end{figure}

To achieve cross-dataset consistency, supporting goal (2), collectors are provided with mock company profiles, including name of the mock company, as well as mock product or service names with associated prices. Collectors are also provided with a schema of intents and associated slots. Some flexibility is allowed in the slot schema to reflect real world situations where customers may not have all requested information on hand. Examples of company profiles and intent schemas are provided in Appendix Tables~\ref{Tab:companyprofile} and \ref{Tab:sampleschema}.
For each conversation, we sample a set of minimum discourse complexity types. For example, one conversation could be assigned the target complexities of ChitChat, FollowUpQuestion, MultiElicit, and SlotLookup. Scenarios eliciting each of these complexity types are generated and provided to the participants.

For \oursspoken, one participant plays the part of the customer service representative (``agent''), and one participant plays the part of the customer (``customer''). The participants are recorded as they play-act the scenarios described on their instruction sheets from the same room. These audio recordings are then transcribed and annotated for actual complexities.

Given the time, cost, and complexity involved for the creation of the \oursspoken{} datasets, as an alternative approach, we apply the \oursself{} method. For \oursself{}, participants write self-dialogues as if they were spoken out loud. This method has the benefit of (1) not needing to be transcribed, and (2) requiring only one participant to create each conversation and therefore not requiring scheduling to match participants together. However, these rely on an understanding by the participants of the distinction between spoken and written modality data.
The \oursself{} method follows a similar set-up as the \oursspoken{} method, except in addition to being provided with a set of target discourse complexities, participants are also provided with a set of \textbf{spoken form complexities}.
While discourse complexities target discourse-related phenomena, spoken form complexities consist of phenomena specifically observed in spoken form speech. For this complexity type, we include phenomena such as hesitations or fillers (`um'), rambling, spelling, and backchanneling (`uh huh go on'). A list of these complexities along with target percentages is provided in Figure~\ref{Fig:spoken}, and further examples are provided in Appendix Table~\ref{Tab:spokenexamples}.

\subsection{Annotations}
\label{Sec:Annotations}

One goal of collecting realistic dialogues is to facilitate the development and evaluation of tools for building task-oriented dialogue systems from H2H conversations. To this end, we perform two types of annotations on a subset of \ours: Dialogue Act (DA) annotations and Intent Classification and Slot Labeling (IC/SL) annotations.

IC/SL annotations are intended to label intents and slots, two key elements of many TOD systems.
An intent is broadly a customer goal, and a slot is a smaller piece of information related to that goal.
We use an open labelset, asking annotators to come up with specific labels for each intent and slot, such as ``BookFlight'' and ``PreferredAirline'' as opposed to simply ``Intent'' and ``Slot''. 
Annotators are instructed to label the same intent no more than once per conversation. 
For slots, we use the principle of labeling the smallest complete grammatical constituent that communicates the necessary information.

Our DA annotations are intended to identify utterances that move the dialog towards the customer's goal.
TOD systems often support only a small set of dialogue acts that capture supported user and agent actions.
For the agent, these may include eliciting the user's intent or asking for slot values associated with that intent (\da{ElicitIntent} and \da{ElicitSlot} respectively).
For the user, such acts may include informing the agent of a new intent or providing relevant details for resolving their request (\da{InformIntent} and \da{InformSlot} respectively).
Such acts provide a limited view of the actions taken by speakers in natural conversations, but do provide a way to identify and categorize automatable interactions in natural conversations. 
Table~\ref{tab:example_conversation_i} provides an example conversation annotated with intents, slots, and dialogue acts.

\section{Dataset Analysis}
\label{Sec:analysis}
To better motivate \ours{} as a proxy for natural, spoken form customer service conversations from multiple domains with a diverse set of intents, we compare with real conversations from commercial datasets comprising 5 call centers for retail and finance-related businesses (henceforth \real{}).
All datasets in \real{} consist of manually-transcribed conversations between human agents and customers in live phone conversations where all personally-identifiable information has been pre-redacted.
We restrict our analysis to datasets with primarily customer-initiated two-party dialogues.

\begin{table*}[tbh]
\centering
\begin{tabular}{rlcccc}
\toprule
\multicolumn{2}{r}{\textbf{Collection}} & \textbf{\# Dialogues} & \textbf{\# Turns/Conv} & \textbf{\# Words/Turn} & \textbf{MTLD} \\
\midrule
\multirow{4}{*}{\textbf{TOD}} & \multidogo{} & 86,719 & \std{15.9}{4.4} & \std{11.9}{12.0} & \std{46.6}{11.5} \\
& \multiwoz{} & 10,437 & \std{13.7}{5.2} & \std{15.4}{7.4} & \std{43.0}{9.5} \\
& \sgd{} & 22,825 & \std{20.3}{7.2} & \std{11.7}{7.2} & \std{38.3}{9.0} \\
& \tmself{} & 7,708 & \std{22.0}{2.8} & \std{10.3}{7.6} & \std{45.9}{12.2} \\
\midrule
\multirow{1}{*}{\textbf{\oursself{}}} & \oursinsuranceself{} & 954 & \std{70.6}{19.2} & \std{12.3}{9.6} & \std{45.4}{10.7} \\
\multirow{4}{*}{\textbf{\oursspoken{}}} & \oursbanking{} & 980 & \std{59.6}{23.1} & \std{18.2}{19.4} & \std{37.4}{6.7} \\
& \oursfinance{} & 3,000 & \std{65.6}{22.4} & \std{16.3}{19.4} & \std{36.9}{6.0} \\
& \ourshealth{} & 1,000 & \std{67.0}{24.8} & \std{16.3}{20.3} & \std{34.7}{7.4} \\
& \ourstravel{} & 1,000 & \std{72.1}{24.7} & \std{16.2}{20.0} & \std{34.7}{7.1} \\
\midrule
\multirow{5}{*}{\textbf{\textbf{\real{}}}} & \realretaila{} & 4,500 & \std{80.3}{41.7} & \std{16.6}{14.9} & \std{30.8}{4.8} \\
& \realretailb{} & 1,400 & \std{52.8}{37.6} & \std{17.9}{14.7} & \std{32.8}{5.9} \\
& \realretailc{} & 4,500 & \std{100.1}{69.9} & \std{14.5}{12.7} & \std{33.6}{5.5} \\
& \realfinancea{} & 1,300 & \std{61.7}{29.1} & \std{17.6}{15.0} & \std{38.4}{6.9} \\
& \realfinanceb{} & 1,700 & \std{69.7}{43.8} & \std{16.1}{13.5} & \std{38.0}{5.9} \\
\bottomrule
\end{tabular}
\caption{
Comparison of dialogue datasets and corresponding high-level data characteristics.
Task-oriented dialogue (TOD) datasets include \multidogo{}~\cite{peskov-etal-2019-multi}, \multiwoz{}~\cite{budzianowski-etal-2018-multiwoz}, \sgd{}~\cite{rastogi2020towards}, and \tmself{}~\cite{byrne-etal-2019-taskmaster}. We compare datasets collected from our two methodologies (\oursself{} and \oursspoken{)}, as well as 5 call center datasets (\real{}). MTLD is a lexical diversity measure~\cite{mccarthy2010mtld} computed at the conversation level.
}
\label{Tab:summary_stats}
\end{table*}
As shown in Table~\ref{Tab:summary_stats}, one surface-level distinction from publicly available TOD datasets is the average number of turns per conversation.
Compared to \multidogo{}, \sgd{}, \multiwoz{} and \tmself{}, \real{} has considerably longer conversations (over 70 turns per conversation on average, vs. 22 for \tmself{}).
Furthermore, each turn has more words per turn, suggesting increasing complexity in spoken H2H dialogues.
\ours{} closely matches \real{} in terms of conversation and turn lengths.

\subsection{Intents and Slots}

\begin{table*}[tbh]
\centering
\addtolength{\tabcolsep}{-3pt}
\begin{tabular}{lccccc}
\toprule
\textbf{Collection} & \textbf{\multidogo{}} & \textbf{\sgd{}} & \textbf{\oursself{}} & \textbf{\oursspoken{}} & \textbf{\real{}} \\
\midrule
Intents & 6.8 & 46.0 & 61.0 & 94.0 & 64.0 \\
Turns/Intent & 473.1 & 1,235.5 & 42.5 & 27.3 & 47.4 \\
Intent Turn Len. & 8.2 & 14.1 & 17.8 & 29.8 & 23.9 \\
Intents/Conv. & 1.3 & 2.5 & 2.7 & 2.4 & 3.3 \\
Sem. Diversity & 21.3 & 28.6 & 34.5 & 31.1 & 43.5 \\
\midrule
Slots & 13.8 & 112.0 & 246.0 & 238.0 & 150.0 \\
Slots/Conv. & 3.7 & 5.8 & 13.5 & 17.5 & 9.3 \\
Slot 2-Gram \% & 2.8 & 1.9 & 16.3 & 12.4 & 24.2 \\
Slot 3-Gram \% & 10.9 & 12.4 & 35.2 & 31.0 & 57.7 \\
\bottomrule
\end{tabular}
\caption{
Intent and slot annotation statistics for \real{}, \ours{}, \multidogo{}, and \sgd{}.
\textit{Sem. Diversity} is an sentence embedding-based metric for measuring intent-level diversity reported in ~\citet{casanueva-etal-2022-nlu}. \textit{Slot n-gram \%} indicates the ratio of unique sequences of slot annotations in conversations to the total number of such sequences.
} 
\label{Tab:intent_slot_stats}
\end{table*}

Table~\ref{Tab:intent_slot_stats} provides a comparison of intent and slot annotations between existing synthetic datasets, \ours{} and \real{}.
Datasets in \real{} contain considerably more intents and slots for a particular domain than existing TOD datasets like \multidogo.
Turns containing intents are longer for both \ours{} and \real{} than \sgd{} and \multidogo{}.

\begin{figure}[tbh]
\includegraphics[width=0.5\textwidth]{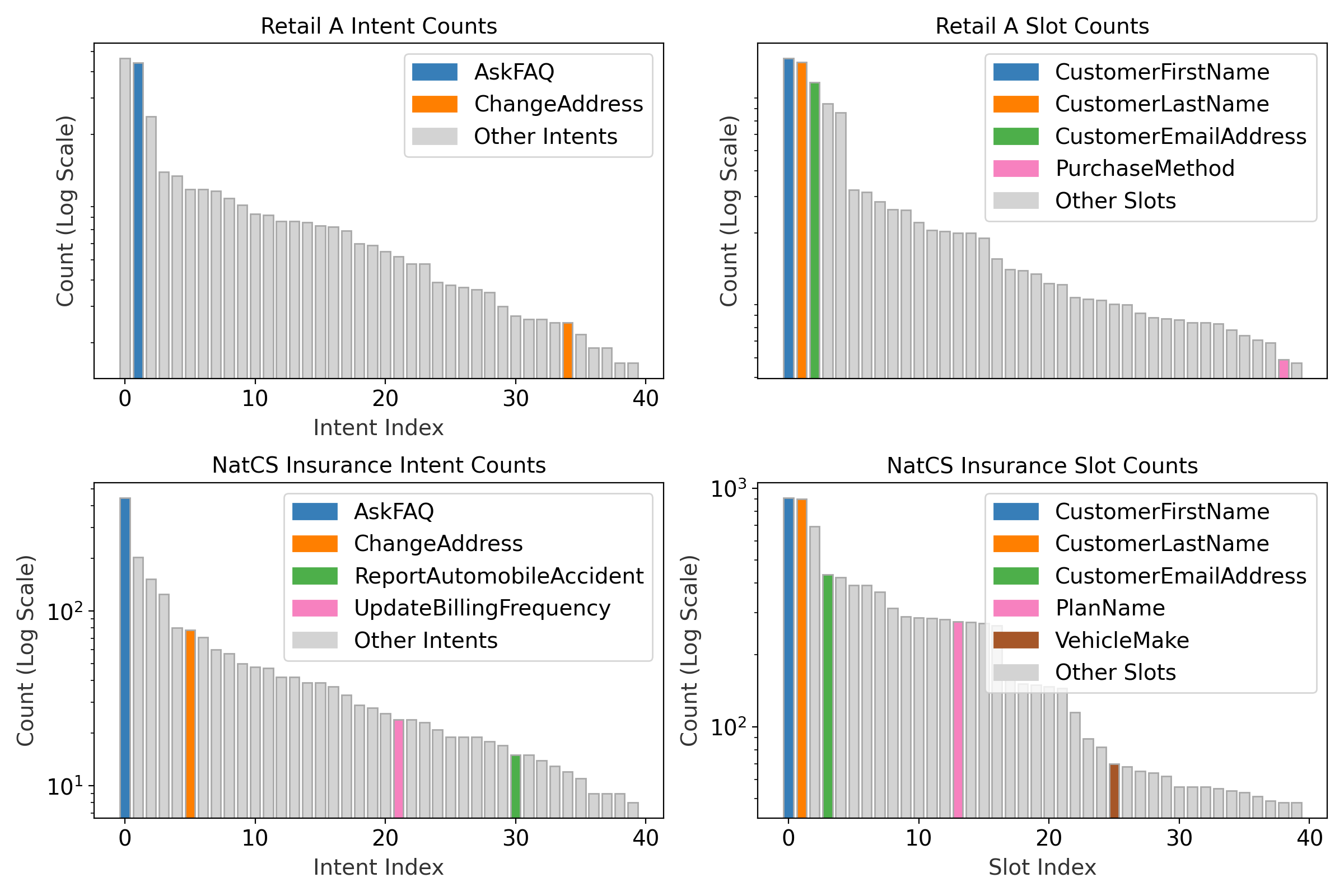}
\centering
\caption{Intent and slot counts (logarithmic scale) for \realretaila{} and \ours{} \oursinsuranceself{}. We observe intents and slots in real data follow a Zipfian distribution in both \real{} and \ours{}.}
\label{Fig:intent_slot_distribution}
\end{figure}

Figure~\ref{Fig:intent_slot_distribution} compares the intent and slot distributions between \realretaila{} and \oursself{} \oursinsuranceself{}, indicating that both have skewed, long-tailed distributions of intents/slots, a product of the open intent/slot schemas used in \ours{}.

\subsection{Diversity Metrics}
As we expect conversations in \real{} to be less homogeneous than synthetic dialogue datasets, we compute automatic metrics to measure multiple aspects of diversity and compare with \ours.

\paragraph{Conversational Diversity}
In Table~\ref{Tab:intent_slot_stats}, we examine the diversity of conversation flows as measured by the ratio of unique sequences of slots informed by the customer to the total number of sequences (e.g. slot \textit{bigrams} or \textit{trigrams}).
In \sgd{}, which constructs dialogue templates using a simulator, we observe a much lower percentage of unique n-grams than in \real{}, despite \sgd{} containing dialogues spanning multiple domains and services.
On the other hand, while \ours{} has lower slot n-gram diversity than \real{}, both collection types have substantially higher slot n-gram diversity scores than both \multidogo{} or \sgd{}.

\begin{table}[tbh]
\centering
\begin{tabular}{lcc}
\toprule
\textbf{Collection} & \textbf{PPL} & \textbf{PPL (ZS)} \\
\midrule
\multidogo{} & 4.9 & 10.7 \\
\multiwoz{} & 6.8 & 14.6 \\
\sgd{} & 7.0 & 10.1 \\
\tmself{} & 10.4 & 17.0 \\
\midrule
\oursself{} & 11.6 & 15.7 \\
\oursspoken{} & 11.7 & 16.7 \\
\real{} & 15.5 & 22.4 \\
\bottomrule
\end{tabular}
\caption{Perplexity of GPT-Neo 125M on datasets in both fine-tuning (PPL) and zero-shot (ZS PPL) settings.}
\label{Tab:perplexity}
\end{table}

The average perplexity of a language model provides one indication of the difficulty of modeling the dialogue in a given dataset~\cite{byrne-etal-2019-taskmaster}.
High perplexity indicates higher difficulty, while low perplexity can indicate more uniform or predictable datasets.
We compare between fine-tuning and zero-shot language modeling settings using \textsc{GPT-neo}~\cite{gpt-neo}.
Zero-shot evaluation gives an indication of how compatible the dataset is with the pre-trained model without any fine-tuning.
Section~\ref{sec:Appendix:ExperimentalSettings} provides details on the fine-tuning procedure.

As shown in Table~\ref{Tab:perplexity}, we observe high perplexity on real data and low perplexity on synthetic datasets like \sgd{} and \multidogo{}.
\ours{} has lower perplexity than \real{}, but considerably higher perplexity than \multidogo{}, \multiwoz{}, and \sgd{}.
Interestingly, there is a wider range of perplexities across real datasets, while most existing TOD datasets have a perplexity of 10 or less.

\paragraph{Intent Diversity}
We also investigate the semantic diversity of intent turns (\semdiv{}) following~\citet{casanueva-etal-2022-nlu}.
Details of this calculation are provided in Section~\ref{sec:Appendix:ExperimentalSettings}.
As shown in Table~\ref{Tab:intent_slot_stats}, we observe the highest \semdiv{} for \real{}, but \ours{} has considerably higher \semdiv{} than pre-existing synthetic datasets, indicating greater potential modeling challenges.
We also compare the semantic of diversity of \ours{} with other datasets for specific aligned intents, like \textit{CheckBalance} in Appendix Table~\ref{Tab:semantic-diversity}, also observing higher semantic diversity as compared to pre-existing intent classification benchmarks.
Further investigation into the lexical diversity of intents is provided in Section~\ref{sec:Appendix:ExperimentalSettings}.

\paragraph{\Ourdas{} Distribution}
\begin{table*}[tbh]
\centering
\addtolength{\tabcolsep}{-3pt}
\begin{tabular}{lccccccc}
\toprule
\textbf{Collection} & \textbf{Total (\%)} & \textbf{InformIntent} & \textbf{ElicitSlot} & \textbf{ElicitIntent} & \textbf{InformSlot} & \textbf{ConfirmSlot} \\
\midrule
\multidogo{}\textsuperscript{*} & 55.4 & \std{1.1}{0.8} & \std{2.7}{1.3} & \std{2.0}{0.8} & \std{3.1}{1.4} & \std{0.7}{0.9} \\
\multirow[t]{2}{*}{\multiwoz{}}\textsuperscript{*} & 70.5 & \std{3.5}{1.9} & \std{1.8}{1.4} & \std{2.1}{1.4} & \std{3.4}{2.1} & \std{2.0}{1.4} \\
\tmself{}\textsuperscript{*} & 52.9 & \std{2.4}{1.7} & \std{3.3}{1.8} & \std{1.0}{1.1} & \std{4.1}{1.7} & \std{2.4}{1.7} \\
\sgd{}  & 60.6 & \std{4.5}{2.3} & \std{2.8}{1.7} & \std{1.3}{1.1} & \std{5.3}{2.3} & \std{6.2}{4.1} \\
\midrule
\oursself{} & 42.9 & \std{4.5}{2.8} & \std{9.2}{3.6} & \std{3.1}{1.2} & \std{11.8}{4.7} & \std{8.2}{5.6} \\
\oursspoken{} & 44.7 & \std{3.5}{2.1} & \std{9.8}{5.2} & \std{3.3}{1.5} & \std{13.5}{6.8} & \std{7.4}{5.9} \\
\real{} & 27.7 & \std{4.3}{3.1} & \std{5.2}{3.9} & \std{2.0}{1.1} & \std{7.3}{5.4} & \std{4.3}{4.9} \\
\bottomrule
\end{tabular}
\caption{Percentage of turns containing \ourdas{} in synthetic and real dialogue datasets. \textsuperscript{*}Indicates distribution estimated from automatic predictions.
A lower percentage of task-oriented turns is observed in \ours{} and \real{} than previous task-oriented dialogue datasets.
}
\label{Tab:da_distribution}
\end{table*}

To better understand the characteristics of typical H2H customer service dialogues in call centers, we annotate a subset of real conversations with the ``task-oriented'' dialogue acts described in Section~\ref{Sec:Annotations}.
Because \ourdas{} consist of a small set of intent and slot-related functions commonly employed in automated TOD systems, we do not expect the labels to have high coverage in real conversations that do not revolve around a fixed set of intents and slots.
However, they aim to provide a mechanism for aligning turns from natural conversations onto these automatable TOD constructs.

Table~\ref{Tab:da_distribution} compares the percentage of turns labeled with \ourdas{} across multiple datasets along with the average counts of each label per dialogue.
As expected, compared to synthetic data, dialogues in \real{} have fewer turns labeled with \ourdas{} (27.7\%).
More reflective of \real{} in this regard, in \oursspoken{} and \oursself{}, we observe that more than half of the turns in each conversation are not labeled, despite having higher total counts of dialogue acts per conversation as compared to pre-existing datasets.

\subsection{Human Evaluation}

We also perform a human evaluation comparing \multidogo{}, \sgd{}, and \real{} datasets against both \oursself{} and \oursspoken{}.
Rather than compare complete dialogues, because of the large disparity in conversation lengths, we restrict evaluation only to snippets including the first 5 turns after an intent is stated (including the turn containing the intent). 
Conversation snippets are graded on a scale of 1 to 5 along multiple dimensions, including \textbf{realism} (believability of the dialogue), \textbf{concision} (conciseness of customer, lack of verbosity), and \textbf{spoken-likeness} (possibility of being part of a spoken conversation). See Section~\ref{sec:Appendix:HumanEvaluation} Figure~\ref{Fig:guidelines} for explicit definitions provided to graders.

The evaluation is conducted by 6 dialogue systems researchers, with each grader rating 50 randomly-selected conversations.
As indicated by results in Table~\ref{Tab:human_evaluation}, conversations from \real{} observe high values for both realism and spoken-likeness (4.71 and 4.95 respectively), with lower values for concision, indicating greater customer verbosity in real dialogues.
The results also indicate that although expert human graders can still differentiate \ours{} from real conversations, \ours{} is graded as significantly more realistic and indicative of spoken modality than \sgd{} and \multidogo{} (two-tailed T-test with $p<0.005$).

\begin{table}[tbh]
\centering
\addtolength{\tabcolsep}{-3pt}
\begin{tabular}{llll}
\toprule
\textbf{Dataset} & \textbf{Realism} & \textbf{Concision} & \textbf{Spokenlike} \\
\midrule
\multidogo{} & 2.85 & 4.21 & 3.18 \\
\sgd{} & 3.36\textsuperscript{*} & 4.66\textsuperscript{**} & 3.18 \\
\midrule
\oursself{} & 4.06\textsuperscript{**} & 3.88\textsuperscript{**} & 4.17\textsuperscript{**} \\
\oursspoken{} & 4.34 & 3.58 & 4.53 \\
\real{} & 4.71\textsuperscript{*} & 2.87\textsuperscript{**} & 4.95\textsuperscript{**} \\
\bottomrule
\end{tabular}
\caption{
Human ratings comparing conversation snippets between \real{}, \ours{} and other task-oriented dialogue datasets along multiple dimensions (see Section~\ref{sec:Appendix:HumanEvaluation}). Significance is based two-tailed t-tests ($\textsuperscript{*}p < 0.05$, $\textsuperscript{**}p < 0.005$), comparing samples in adjacent rows. For \multidogo{}, we observe no significant difference for Concision ($p = 0.0522$) when comparing with \oursself{}, but do for \oursspoken{} ($p = 0.0013$). 
}
\label{Tab:human_evaluation}
\end{table}

\section{Applications}
\label{Sec:applications}
In this section, we investigate two potential applications of \ours{} as a resource for building and evaluating systems related to human-to-human (H2H) customer service interactions.
One goal of \ours{} is to encourage research in the under-explored space of H2H task-oriented interactions, so these are intended to serve as motivating examples rather than prescribed uses.

\subsection{\ours{} as Training Data}
One goal of this work is to accelerate the development of dialogue systems based on H2H conversations.
While most existing work in intent induction assumes that customer turns corresponding to requests have already been identified, \ourdas{} provide a mechanism to map turns onto TOD constructs like intents.

To validate the usefulness of dialogue act annotations in \ours{}, we compare the cross-dataset generalization of dialogue act classifiers trained on annotations in \ours{} against that of SGD, a large multi-domain corpus of task-oriented dialogues, evaluating on real conversations between human agents and customers.

\begin{table}[tbh]
\centering
\addtolength{\tabcolsep}{-3pt}
\begin{tabular}{lccc}
\toprule
\textbf{Train} & $\text{\textbf{P}} \uparrow$  & $\text{\textbf{R}} \uparrow$ & $\text{\textbf{F}}_1 \uparrow$  \\
\midrule
\sgd{} & \std{40.8}{1.6} & \std{26.1}{3.1} & \std{31.6}{2.7} \\
\ours & \std{64.6}{2.3} & \std{46.8}{3.1} & \std{54.1}{1.8} \\
\realood{} & \std{63.3}{3.2} & \std{57.2}{2.9} & \std{59.6}{0.6} \\
\realid{} & \std{67.1}{2.4} & \std{61.9}{2.5} & \std{64.2}{0.5} \\
\bottomrule
\end{tabular}
\caption{
Comparison of dialogue act classifier performance on real datasets trained on \sgd{}, \ours, and in-domain (\realid{}) vs. out-of-domain (\realood{}) real data. Training on \ours{} achieves comparable precision to \realood{}.
}
\label{Tab:da_classifier_real}
\end{table}

We fine-tune \textsc{RoBERTa-base} using per-label binary cross entropy losses to support multiple labels per sentence.
Fine-tuning details are provided in Section~\ref{sec:Appendix:DialogueActClassifierTraining}.
We compare the dialogue act classification performance on real data when training on \sgd{}, \ours{}, and in-domain vs. out-of-domain real data.
As shown in Table~\ref{Tab:da_classifier_real}, a DA classifier trained on \ours{} performs significantly better on real data than a classifier trained on SGD.
Performance still lags behind that of training on real data, but with \ours{}, the gap is closed considerably.
In Section~\ref{sec:Appendix:DialogueActDomainGeneralization}, we also show that the dialogue act annotations in \ours{} are able to generalize to new domains.

\subsection{Intent Clustering with Noise}
Recent work indicates growing interest in applications that can accelerate the development of these systems by automatically inducing TOD constructs such as intents and slots from customer support interactions~\cite{yu-etal-2022-unsupervised, shen-etal-2021-semi, kumar-etal-2022-intent, perkins-yang-2019-dialog, chatterjee-sengupta-2020-intent}.
To further motivate \ours{} as a realistic test bed for applications that learn from natural conversations, we demonstrate how it can serve as a benchmark for  unsupervised intent clustering tasks.

In a realistic setting, turns in conversations containing intents will not be provided in advance.
We thus compare three settings: 1) using the first customer turn in each conversation 2) using turns predicted as having intents with a dialogue act classifier and 3) using turns labeled with intents (\textit{gold} dialogue acts).

Utterances are encoded using a sentence embedding model from the \textsc{SentenceTransformers} library~\cite{reimers-gurevych-2019-sentence}, \textsc{all-mpnet-base-v2}.
We use k-means clustering with the number of clusters set to the number of reference intents.
To assign cluster labels to all gold input turns, we use label propagation by training a logistic regression classifier using \textsc{all-mpnet-base-v2} embeddings as static features on inputs assigned cluster labels (such as first turns), then apply the classifier to the missing turns to get predicted cluster labels.

\begin{table}[tb]
\centering
\begin{tabular}{lcccc}
\toprule
 \textbf{Turns} & \textbf{NMI} & \textbf{ACC} & \textbf{Purity} & \textbf{Inv. Purity} \\
\midrule
 First & 53.3 & 44.8 & 50.2 & 58.9 \\
Pred. & 56.7 & 49.1 & 55.7 & 59.9 \\
 Gold & 61.6 & 49.4 & 63.9 & 55.0 \\
\bottomrule
\end{tabular}
\caption{Intent clustering performance comparing different intent turn identification strategies (First, for first customer turns in each conversation, and Pred. for predicted dialogue acts) with using gold intent turns (Gold), averaged over Finance, Insurance, and Banking datasets in \ours{}.}
\label{Tab:intent_clustering}
\end{table}

The results, shown in Table~\ref{Tab:intent_clustering}, demonstrate that using automatically predicted turns leads to a drop in purity and NMI.
The drop in purity is attributable to irrelevant, non-intentful turns being clustered together with relevant intents, a potentially costly error in real-world settings that is not typically reflected in intent clustering evaluation.
\section{Conclusions}
\label{Sec:conclusion}
We present \ours{}, a corpus of realistic spoken human-to-human customer service conversations.
The collection of \ours{} is complexity-driven and domain-restricted, resulting in a dataset that better approximates real conversations than pre-existing task-oriented dialogue datasets along a variety of both automated and human-rated metrics.
We demonstrate two potential downstream applications of \ours{}, showing that training using \ours{} results in better performance with real test data compared to training using other publicly-available goal-oriented datasets, and that \ours{} can provide a new challenging benchmark for realistic evaluation of intent induction.

We hope that \ours{} will help facilitate open research in applications based on customer support conversations previously accessible mainly in industry settings by providing a more realistic annotated dataset.
In future work, we hope to expand on annotations in \ours{} to support more tasks such as call summarization, response selection, and generation.

\section*{Limitations}
\ours{} is partially annotated with dialogue acts, intents, and slots, which are annotated independently from the initial collection of the conversations.
While decoupling annotations from collection was intended to facilitate natural and diverse dialogues, the methodology is more time-consuming and expensive than previous approaches that use pre-structured conversation templates to avoid the need for manual annotation.
In particular, \oursspoken{} requires multiple participants engaging in synchronous conversations, followed by independent manual transcriptions and annotations, making the approach particularly time-consuming and difficult to apply for large collections.
Furthermore, this decoupling of annotations from collection has greater potential for annotator disagreement.

While the complexity types and annotations are mostly language-agnostic, \ours{} is restricted to \textsc{en-US} customer-initiated customer service conversations between a single agent and customer in a limited number of domains (multi-party conversations beyond two participants or agent-initiated conversations are not included).
The annotations included are primarily intended for applications related to task-oriented dialogue systems.

Further, we note that \ours{} closes the gap from real conversations along many metrics, but still falls short along some dimensions.
We find that real conversations are more verbose, more believable, and less predictable.
We also note that comparisons in our paper focused on a limited number of task-oriented dialogue datasets with different collection approaches, and did not exhaustively include all pre-existing dialogue datasets for comparison.

\section*{Ethics Statement}
In this paper, we present a new partially-annotated dataset.
In adherence with the ACL code of conduct and recommendations laid out in~\citet{bender-friedman-2018-data} it is appropriate to include a data statement.
Our dataset is completely novel, and was collected specifically to support the development of natural language systems.
Workers who are proficient in the \textsc{en-US} variant of English were hired through a vendor with a competitive hourly rate compared to the industry standard for language consultants.
For \oursspoken{}, these workers spoke to each other and then transcribed the data.
For \oursself{} these workers wrote the conversations.

To annotate the data, we used two pools of annotators.
Both had formal training in linguistics and were proficient in the \textsc{en-US} variant of English. One pool was hired through a vendor with a competitive hourly rate.
The other pool consisted of full-time employees.

\paragraph{Curation Rationale}
Our dataset includes all of the data that was produced by the consultants we hired.
Quality Assurance was done on a subset of this data.
We hope that any concerns would have shown up in this sample.
We annotated a random subset of the full dataset.

\paragraph{Language Variety}
The dataset is \textsc{en-US}. The speakers (or writers) were all fluent speakers of \textsc{en-US}. We did not target a particular sub-type of the \textsc{en-US} language variety.

\paragraph{Speaker Demographics}
We do not have detailed speaker demographics, however, we do have male and female speakers from a variety of age ranges.

\paragraph{Annotator Demographics}
We do not have detailed annotator demographics, however, we do have male and female speakers from a variety of age ranges.
All annotators had at least some formal linguistics training (ranging from a B.A. to a Ph.D.).

\paragraph{Speech Situation}
For \oursspoken{}, speakers were talking in real time on the phone to one another.
It was semi-scripted.
Speakers were not told exactly what to say, but were given some constraints.

\bibliography{anthology,custom}

\begin{thebibliography}{27}
\expandafter\ifx\csname natexlab\endcsname\relax\def\natexlab#1{#1}\fi

\bibitem[{{Amazon Contact Lens}(2023)}]{AmazonContactLens}
{Amazon Contact Lens}. 2023.
\newblock {Contact Lens}.
\newblock \url{https://aws.amazon.com/connect/contact-lens/}.

\bibitem[{Bender and Friedman(2018)}]{bender-friedman-2018-data}
Emily~M. Bender and Batya Friedman. 2018.
\newblock \href {https://doi.org/10.1162/tacl_a_00041} {Data statements for
  natural language processing: Toward mitigating system bias and enabling
  better science}.
\newblock \emph{Transactions of the Association for Computational Linguistics},
  6:587--604.

\bibitem[{Black et~al.(2021)Black, Gao, Wang, Leahy, and Biderman}]{gpt-neo}
Sid Black, Leo Gao, Phil Wang, Connor Leahy, and Stella Biderman. 2021.
\newblock \href {https://doi.org/10.5281/zenodo.5297715} {{GPT-Neo: Large Scale
  Autoregressive Language Modeling with Mesh-Tensorflow}}.

\bibitem[{Budzianowski et~al.(2018)Budzianowski, Wen, Tseng, Casanueva, Ultes,
  Ramadan, and Ga{\v{s}}i{\'c}}]{budzianowski-etal-2018-multiwoz}
Pawe{\l} Budzianowski, Tsung-Hsien Wen, Bo-Hsiang Tseng, I{\~n}igo Casanueva,
  Stefan Ultes, Osman Ramadan, and Milica Ga{\v{s}}i{\'c}. 2018.
\newblock \href {https://doi.org/10.18653/v1/D18-1547} {{M}ulti{WOZ} - a
  large-scale multi-domain {W}izard-of-{O}z dataset for task-oriented dialogue
  modelling}.
\newblock In \emph{Proceedings of the 2018 Conference on Empirical Methods in
  Natural Language Processing}, pages 5016--5026, Brussels, Belgium.
  Association for Computational Linguistics.

\bibitem[{Byrne et~al.(2019)Byrne, Krishnamoorthi, Sankar, Neelakantan,
  Goodrich, Duckworth, Yavuz, Dubey, Kim, and
  Cedilnik}]{byrne-etal-2019-taskmaster}
Bill Byrne, Karthik Krishnamoorthi, Chinnadhurai Sankar, Arvind Neelakantan,
  Ben Goodrich, Daniel Duckworth, Semih Yavuz, Amit Dubey, Kyu-Young Kim, and
  Andy Cedilnik. 2019.
\newblock \href {https://doi.org/10.18653/v1/D19-1459} {Taskmaster-1: Toward a
  realistic and diverse dialog dataset}.
\newblock In \emph{Proceedings of the 2019 Conference on Empirical Methods in
  Natural Language Processing and the 9th International Joint Conference on
  Natural Language Processing (EMNLP-IJCNLP)}, pages 4516--4525, Hong Kong,
  China. Association for Computational Linguistics.

\bibitem[{Casanueva et~al.(2020)Casanueva, Tem{\v{c}}inas, Gerz, Henderson, and
  Vuli{\'c}}]{casanueva-etal-2020-efficient}
I{\~n}igo Casanueva, Tadas Tem{\v{c}}inas, Daniela Gerz, Matthew Henderson, and
  Ivan Vuli{\'c}. 2020.
\newblock \href {https://doi.org/10.18653/v1/2020.nlp4convai-1.5} {Efficient
  intent detection with dual sentence encoders}.
\newblock In \emph{Proceedings of the 2nd Workshop on Natural Language
  Processing for Conversational AI}, pages 38--45, Online. Association for
  Computational Linguistics.

\bibitem[{Casanueva et~al.(2022)Casanueva, Vuli{\'c}, Spithourakis, and
  Budzianowski}]{casanueva-etal-2022-nlu}
Inigo Casanueva, Ivan Vuli{\'c}, Georgios Spithourakis, and Pawe{\l}
  Budzianowski. 2022.
\newblock \href {https://doi.org/10.18653/v1/2022.findings-naacl.154} {{NLU}++:
  A multi-label, slot-rich, generalisable dataset for natural language
  understanding in task-oriented dialogue}.
\newblock In \emph{Findings of the Association for Computational Linguistics:
  NAACL 2022}, pages 1998--2013, Seattle, United States. Association for
  Computational Linguistics.

\bibitem[{Chatterjee and Sengupta(2020)}]{chatterjee-sengupta-2020-intent}
Ajay Chatterjee and Shubhashis Sengupta. 2020.
\newblock \href {https://doi.org/10.18653/v1/2020.coling-main.366} {Intent
  mining from past conversations for conversational agent}.
\newblock In \emph{Proceedings of the 28th International Conference on
  Computational Linguistics}, pages 4140--4152, Barcelona, Spain (Online).
  International Committee on Computational Linguistics.

\bibitem[{Chen et~al.(2021)Chen, Chen, Yang, Lin, and
  Yu}]{chen-etal-2021-action}
Derek Chen, Howard Chen, Yi~Yang, Alexander Lin, and Zhou Yu. 2021.
\newblock \href {https://doi.org/10.18653/v1/2021.naacl-main.239} {Action-based
  conversations dataset: A corpus for building more in-depth task-oriented
  dialogue systems}.
\newblock In \emph{Proceedings of the 2021 Conference of the North American
  Chapter of the Association for Computational Linguistics: Human Language
  Technologies}, pages 3002--3017, Online. Association for Computational
  Linguistics.

\bibitem[{El~Asri et~al.(2017)El~Asri, Schulz, Sharma, Zumer, Harris, Fine,
  Mehrotra, and Suleman}]{el-asri-etal-2017-frames}
Layla El~Asri, Hannes Schulz, Shikhar Sharma, Jeremie Zumer, Justin Harris,
  Emery Fine, Rahul Mehrotra, and Kaheer Suleman. 2017.
\newblock \href {https://doi.org/10.18653/v1/W17-5526} {{F}rames: a corpus for
  adding memory to goal-oriented dialogue systems}.
\newblock In \emph{Proceedings of the 18th Annual {SIG}dial Meeting on
  Discourse and Dialogue}, pages 207--219, Saarbr{\"u}cken, Germany.
  Association for Computational Linguistics.

\bibitem[{Eric et~al.(2017)Eric, Krishnan, Charette, and
  Manning}]{eric-etal-2017-key}
Mihail Eric, Lakshmi Krishnan, Francois Charette, and Christopher~D. Manning.
  2017.
\newblock \href {https://doi.org/10.18653/v1/W17-5506} {Key-value retrieval
  networks for task-oriented dialogue}.
\newblock In \emph{Proceedings of the 18th Annual {SIG}dial Meeting on
  Discourse and Dialogue}, pages 37--49, Saarbr{\"u}cken, Germany. Association
  for Computational Linguistics.

\bibitem[{{Google Contact Center AI}(2023)}]{GoogleContactCenterAI}
{Google Contact Center AI}. 2023.
\newblock {Contact Center}.
\newblock \url{https://cloud.google.com/solutions/contact-center}.

\bibitem[{Hewitt and Beaver(2020)}]{hewitt-beaver-2020-case}
Timothy Hewitt and Ian Beaver. 2020.
\newblock \href {https://aclanthology.org/2020.sigdial-1.11} {A case study of
  user communication styles with customer service agents versus intelligent
  virtual agents}.
\newblock In \emph{Proceedings of the 21th Annual Meeting of the Special
  Interest Group on Discourse and Dialogue}, pages 79--85, 1st virtual meeting.
  Association for Computational Linguistics.

\bibitem[{Kelley(1984)}]{kelley1984iterative}
John~F Kelley. 1984.
\newblock An iterative design methodology for user-friendly natural language
  office information applications.
\newblock \emph{ACM Transactions on Information Systems (TOIS)}, 2(1):26--41.

\bibitem[{Kumar et~al.(2022)Kumar, Patidar, Varshney, Vig, and
  Shroff}]{kumar-etal-2022-intent}
Rajat Kumar, Mayur Patidar, Vaibhav Varshney, Lovekesh Vig, and Gautam Shroff.
  2022.
\newblock \href {https://doi.org/10.18653/v1/2022.naacl-main.134} {Intent
  detection and discovery from user logs via deep semi-supervised contrastive
  clustering}.
\newblock In \emph{Proceedings of the 2022 Conference of the North American
  Chapter of the Association for Computational Linguistics: Human Language
  Technologies}, pages 1836--1853, Seattle, United States. Association for
  Computational Linguistics.

\bibitem[{Larson et~al.(2019)Larson, Mahendran, Peper, Clarke, Lee, Hill,
  Kummerfeld, Leach, Laurenzano, Tang, and Mars}]{larson-etal-2019-evaluation}
Stefan Larson, Anish Mahendran, Joseph~J. Peper, Christopher Clarke, Andrew
  Lee, Parker Hill, Jonathan~K. Kummerfeld, Kevin Leach, Michael~A. Laurenzano,
  Lingjia Tang, and Jason Mars. 2019.
\newblock \href {https://doi.org/10.18653/v1/D19-1131} {An evaluation dataset
  for intent classification and out-of-scope prediction}.
\newblock In \emph{Proceedings of the 2019 Conference on Empirical Methods in
  Natural Language Processing and the 9th International Joint Conference on
  Natural Language Processing (EMNLP-IJCNLP)}, pages 1311--1316, Hong Kong,
  China. Association for Computational Linguistics.

\bibitem[{Liao et~al.(2017)Liao, Srivastava, and Kapanipathi}]{liao2017measure}
Q~Vera Liao, Biplav Srivastava, and Pavan Kapanipathi. 2017.
\newblock A measure for dialog complexity and its application in streamlining
  service operations.
\newblock \emph{arXiv preprint arXiv:1708.04134}.

\bibitem[{Liu et~al.(2019)Liu, Ott, Goyal, Du, Joshi, Chen, Levy, Lewis,
  Zettlemoyer, and Stoyanov}]{liu2019roberta}
Yinhan Liu, Myle Ott, Naman Goyal, Jingfei Du, Mandar Joshi, Danqi Chen, Omer
  Levy, Mike Lewis, Luke Zettlemoyer, and Veselin Stoyanov. 2019.
\newblock Roberta: A robustly optimized bert pretraining approach.
\newblock \emph{arXiv preprint arXiv:1907.11692}.

\bibitem[{McCarthy and Jarvis(2010)}]{mccarthy2010mtld}
Philip~M McCarthy and Scott Jarvis. 2010.
\newblock Mtld, vocd-d, and hd-d: A validation study of sophisticated
  approaches to lexical diversity assessment.
\newblock \emph{Behavior research methods}, 42(2):381--392.

\bibitem[{{Microsoft Digital Contact Center
  Platform}(2023)}]{MicrosoftContactCenter}
{Microsoft Digital Contact Center Platform}. 2023.
\newblock {Microsoft Digital Contact Center Platform}.
\newblock \url{https://www.microsoft.com/en-us/microsoft-cloud/contact-center}.

\bibitem[{Perkins and Yang(2019)}]{perkins-yang-2019-dialog}
Hugh Perkins and Yi~Yang. 2019.
\newblock \href {https://doi.org/10.18653/v1/D19-1413} {Dialog intent induction
  with deep multi-view clustering}.
\newblock In \emph{Proceedings of the 2019 Conference on Empirical Methods in
  Natural Language Processing and the 9th International Joint Conference on
  Natural Language Processing (EMNLP-IJCNLP)}, pages 4016--4025, Hong Kong,
  China. Association for Computational Linguistics.

\bibitem[{Peskov et~al.(2019)Peskov, Clarke, Krone, Fodor, Zhang, Youssef, and
  Diab}]{peskov-etal-2019-multi}
Denis Peskov, Nancy Clarke, Jason Krone, Brigi Fodor, Yi~Zhang, Adel Youssef,
  and Mona Diab. 2019.
\newblock \href {https://doi.org/10.18653/v1/D19-1460} {Multi-domain
  goal-oriented dialogues ({M}ulti{D}o{GO}): Strategies toward curating and
  annotating large scale dialogue data}.
\newblock In \emph{Proceedings of the 2019 Conference on Empirical Methods in
  Natural Language Processing and the 9th International Joint Conference on
  Natural Language Processing (EMNLP-IJCNLP)}, pages 4526--4536, Hong Kong,
  China. Association for Computational Linguistics.

\bibitem[{Rastogi et~al.(2020)Rastogi, Zang, Sunkara, Gupta, and
  Khaitan}]{rastogi2020towards}
Abhinav Rastogi, Xiaoxue Zang, Srinivas Sunkara, Raghav Gupta, and Pranav
  Khaitan. 2020.
\newblock Towards scalable multi-domain conversational agents: The
  schema-guided dialogue dataset.
\newblock \emph{Proceedings of the AAAI Conference on Artificial Intelligence},
  34(05):8689--8696.

\bibitem[{Reimers and Gurevych(2019)}]{reimers-gurevych-2019-sentence}
Nils Reimers and Iryna Gurevych. 2019.
\newblock \href {https://doi.org/10.18653/v1/D19-1410} {Sentence-{BERT}:
  Sentence embeddings using {S}iamese {BERT}-networks}.
\newblock In \emph{Proceedings of the 2019 Conference on Empirical Methods in
  Natural Language Processing and the 9th International Joint Conference on
  Natural Language Processing (EMNLP-IJCNLP)}, pages 3982--3992, Hong Kong,
  China. Association for Computational Linguistics.

\bibitem[{Shen et~al.(2021)Shen, Sun, Zhang, and
  Najmabadi}]{shen-etal-2021-semi}
Xiang Shen, Yinge Sun, Yao Zhang, and Mani Najmabadi. 2021.
\newblock \href {https://doi.org/10.18653/v1/2021.nlp4convai-1.12}
  {Semi-supervised intent discovery with contrastive learning}.
\newblock In \emph{Proceedings of the 3rd Workshop on Natural Language
  Processing for Conversational AI}, pages 120--129, Online. Association for
  Computational Linguistics.

\bibitem[{Wen et~al.(2016)Wen, Vandyke, Mrksic, Gasic, Rojas-Barahona, Su,
  Ultes, and Young}]{wen2016network}
Tsung-Hsien Wen, David Vandyke, Nikola Mrksic, Milica Gasic, Lina~M
  Rojas-Barahona, Pei-Hao Su, Stefan Ultes, and Steve Young. 2016.
\newblock A network-based end-to-end trainable task-oriented dialogue system.
\newblock \emph{arXiv preprint arXiv:1604.04562}.

\bibitem[{Yu et~al.(2022)Yu, Wang, Cao, Shafran, Shafey, and
  Soltau}]{yu-etal-2022-unsupervised}
Dian Yu, Mingqiu Wang, Yuan Cao, Izhak Shafran, Laurent Shafey, and Hagen
  Soltau. 2022.
\newblock \href {https://doi.org/10.18653/v1/2022.naacl-main.86} {Unsupervised
  slot schema induction for task-oriented dialog}.
\newblock In \emph{Proceedings of the 2022 Conference of the North American
  Chapter of the Association for Computational Linguistics: Human Language
  Technologies}, pages 1174--1193, Seattle, United States. Association for
  Computational Linguistics.

\end{thebibliography}
\bibliographystyle{acl_natbib}

\appendix

\section{Appendix}
\label{sec:appendix}

\subsection{Experimental Setting Details}
\label{sec:Appendix:ExperimentalSettings}
In this section, we provide details on the experimental settings used for evaluation of \ours{} and other dialogue datasets.

\paragraph{Perplexity Evaluation}
To evaluate the language modeling (LM) perplexity on each dataset, we compare both a fine-tuning setting as well as a zero-shot setting using \textsc{GPT-neo}~\cite{gpt-neo} as the pre-trained LM.
We fine-tune \textsc{GPT-neo} on each dataset, sampling 4096 blocks of 128 tokens as training data and evaluating on held-out test splits.
Fine-tuning is performed for 6 epochs with a batch size of 64 and learning rate of 5e-5.
Perplexity is computed at the level of bytes using a sliding window of 128 tokens.

\paragraph{Semantic Diversity Evaluation}
Following~\citet{casanueva-etal-2022-nlu}, to compute semantic diversity for a single intent, we (1) compute intent centroids as the average of embeddings for the turns labeled with the intent using the SentenceBERT~\cite{reimers-gurevych-2019-sentence} library with the pre-trained model~\textsc{all-mpnet-base-v2}, then (2) find the average cosine distance between each individual turn and the resulting centroid.
Finally, (3) overall semantic diversity scores (\semdiv{}) in Table~\ref{Tab:intent_slot_stats} are computed as a frequency-weighted average over intent-level scores.

\begin{table*}[tb!]
\centering
\begin{tabular}{lccccc}
\toprule
 \textbf{Intent} & \textbf{\ours{}} & \textbf{MultiDoGO} & \textbf{CLINC150} & \textbf{BANK77} & \textbf{SGD} \\
\midrule
 CheckBalance & \textbf{31.9} & 17.9 & 27.8 & \_ & 23.1 \\
 MakeTransfer & \textbf{34.3} & 24.2 & 29.5 & \_ & 25.9 \\
ReportLostStolenCard & \textbf{29.0} & 18.6 & 16.2 & 18.4 & \_ \\
 DisputeCharge & \textbf{35.3} & 23.7 & 26.1 & \_ & \_ \\
 OrderChecks & \textbf{31.8} & 21.5 & 19.0 & \_ & \_ \\
 CloseBankAccount & \textbf{26.4} & 17.6 & \_ & 20.1 & \_ \\
 UpdateStreetAddress & \textbf{31.4} & 17.5 & \_ & 28.6 & \_ \\
 ChangePin & \textbf{27.4} & \_ & 20.3 & 19.7 & \_ \\
\bottomrule
\end{tabular}
\caption{Comparing semantic diversity~\cite{casanueva-etal-2022-nlu} for various aligned intents across MultiDoGO~\cite{peskov-etal-2019-multi}, CLINC150~\cite{larson-etal-2019-evaluation}, BANK77~\cite{casanueva-etal-2020-efficient}, and SGD~\cite{rastogi2020towards}.}
\label{Tab:semantic-diversity}
\end{table*}
Table~\ref{Tab:semantic-diversity} shows the semantic diversity scores~\cite{casanueva-etal-2022-nlu} for different intents aligned across several datasets.

\paragraph{Lexical Diversity of Intents}
\begin{table*}[tbh]
\centering
\addtolength{\tabcolsep}{-3pt}
\begin{tabular}{lccccc}
\toprule
\textbf{Collection} & \textbf{\multidogo{}} & \textbf{\sgd{}} & \textbf{\oursself{}} & \textbf{\oursspoken{}} & \textbf{\real{}} \\
\midrule
1-Gram \ittr{} & 7.2 & 2.5 & 16.7 & 17.0 & 16.0 \\
2-Gram \ittr{} & 21.0 & 13.7 & 48.1 & 53.1 & 56.5 \\
3-Gram \ittr{} & 33.0 & 28.6 & 66.7 & 76.6 & 83.2 \\
\bottomrule
\end{tabular}
\caption{
\ittr{} provides intent-level type-token-ratios after removing names and numbers.
Type-token-ratios are computed as 1-grams, 2-grams, and 3-grams, for which we observe consistently higher diversity for \ours{} and \real{} than \multidogo{} and \sgd{}.
} 
\label{Tab:intent_slot_stats_overflow}
\end{table*}
As an additional indicator of intent-level diversity, we measure the frequency-weighted average of type token ratios for utterances within each intent (\ittr{}).
To account for the redaction of names and numbers in real data, we perform a similar redaction step on all datasets, automatically converting names and numbers to a single \textsc{PII} token with regular expressions and a named entity tagger before computing TTR.
Shown in Table~\ref{Tab:intent_slot_stats_overflow}, we observe similar \ittr{} between \ours{} and \real{}, while the pre-existing synthetic datasets lag behind considerably.

\subsection{Human Evaluation Details}
\label{sec:Appendix:HumanEvaluation}
\begin{figure*}[tbh]
\begin{framed}

Read each conversation excerpt, then fill in values for each of the below dimensions with ratings between 1 and 5.  \\

\textbf{Realism} – Is the customer making a realistic request? Is it believable that the conversation took place in a real setting?
Is the customer's request or problem something that you imagine customers can encounter, or is it something incredibly unlikely or silly? Does the agent respond in a professional manner? Is it believable that the conversation took place in real life?

\begin{itemize}
\item 5 (realistic) means that the conversation is realistic and could have taken place in a real life setting.
\item 1 (unrealistic) means that it is highly unlikely that the conversation could have taken place in a real life setting.
\end{itemize}

\textbf{Concision} – Does the customer provide only the necessary details for resolving their request?
Does the customer avoid providing long responses or extra details that aren't 100\% relevant to the conversation or for resolving their request?

\begin{itemize}
\item 5 (concise) means that the customer was consistently concise and clear with their responses to the agent.
\item  1 (not concise) means that the customer provides multiple extra unnecessary details to the agent and provides long responses.
\end{itemize}

\textbf{Spoken-like} – Does the excerpt appear to be from a spoken (as opposed to written) conversation?
Are there signals that indicate that the conversation was originally spoken and then transcribed (as opposed to occurring over a chatroom or messaging platform)?

\begin{itemize}
    \item 5 (spoken) means that it is extremely likely that the conversation was originally spoken.
    \item 1 (written) means that it is unlikely that the conversation was originally spoken (and was probably written as a chat conversation instead).
\end{itemize}
\end{framed}
\caption{Guidelines used in human evaluation.}
\label{Fig:guidelines}
\end{figure*}

Guidelines provided to human graders are provided in Figure~\ref{Fig:guidelines}.
\textbf{Realism} measures the overall believability that the conversation could have taken place in a real scenario (which penalizes unlikely, silly utterances from the customer or unprofessional behavior from the agent).
\textbf{Concision} measures how concise the customer responses are, with lower scores for lengthy utterances containing details that are unnecessary for resolving their request.
\textbf{Spoken-like} measures the likelihood that the conversation was originally spoken in a phone conversation (as opposed to being written from a chatroom or messaging platform).

50 of the 300 conversation snippets evaluated were graded by pairs of annotators (5 pairs total). For these, we observed a Krippendorff's alpha of 0.52, 0.59, and 0.37 respectively for Spoken-like, Realism, and Concision respectively.

\subsection{Dialogue Act Classifier Training Details}
\label{sec:Appendix:DialogueActClassifierTraining}
The dialogue act classifier is implemented by fine-tuning \textsc{RoBERTa-base}~\cite{liu2019roberta} using per-label binary cross entropy losses to support multiple labels per sentence.
To encode dialogue context, we append the three previous sentences to the current turn with a separator token ($ \left[SEP\right] $), adding a speaker label to each sentence and using padding tokens at the beginning of the conversation.
Fine-tuning is performed for 6 epochs with a batch size of 16, using AdamW with a learning rate of 2e-5.
Dataset-level dialogue act classifier performance is provided in Table~\ref{Tab:da_classifier}.

\begin{table*}[tbh]
\centering
\begin{tabular}{lcccc}
\toprule
\textbf{Test}  & \textbf{Train} & $\text{\textbf{P}} \uparrow$  & $\text{\textbf{R}} \uparrow$ & $\text{\textbf{F}}_1 \uparrow$  \\
\midrule
\multirow[t]{2}{*}{\realfinancea{}} & \sgd{} & \std{41.0}{0.7} & \std{30.9}{2.3} & \std{35.2}{1.7} \\
 & \ours & \std{67.1}{1.5} & \std{45.7}{3.3} & \std{54.3}{2.8} \\
\multirow[t]{2}{*}{\realfinanceb{}} & \sgd{} & \std{42.5}{1.1} & \std{29.8}{2.9} & \std{35.0}{2.1} \\
 & \ours & \std{59.5}{1.2} & \std{43.5}{2.5} & \std{50.3}{2.1} \\
\multirow[t]{2}{*}{\realretaila{}} & \sgd{} & \std{42.5}{2.6} & \std{20.3}{3.6} & \std{27.4}{3.7} \\
 & \ours & \std{65.4}{4.4} & \std{47.3}{3.8} & \std{54.7}{1.2} \\
\multirow[t]{2}{*}{\realretailb{}} & \sgd{} & \std{37.2}{2.1} & \std{23.3}{3.5} & \std{28.6}{3.3} \\
 & \ours & \std{66.1}{2.3} & \std{50.5}{2.9} & \std{57.1}{1.1} \\
 \midrule
\multirow[c]{4}{*}{Average} & \sgd{} & \std{40.8}{1.6} & \std{26.1}{3.1} & \std{31.6}{2.7} \\
 & \ours & \std{64.6}{2.3} & \std{46.8}{3.1} & \std{54.1}{1.8} \\
 & \realood{} & \std{63.3}{3.2} & \std{57.2}{2.9} & \std{59.6}{0.6} \\
 & \realid{} & \std{67.1}{2.4} & \std{61.9}{2.5} & \std{64.2}{0.5} \\
\bottomrule
\end{tabular}
\caption{
Comparison of dialogue act classifier performance on real datasets trained on \sgd{}, \ours, and in-domain (\realid{}) vs. out-of-domain (\realood{}) real data. Training on \ours{} achieves comparable precision to \realood{}.
}
\label{Tab:da_classifier}
\end{table*}

\subsection{Dialogue Act Cross-Domain Generalization}
\label{sec:Appendix:DialogueActDomainGeneralization}
\begin{table*}[tbh]
\centering
\begin{tabular}{lcccc}
\toprule
\textbf{Test}  & \textbf{Train} & $\text{\textbf{P}} \uparrow$  & $\text{\textbf{R}} \uparrow$ & $\text{\textbf{F}}_1 \uparrow$  \\
\midrule
\multirow[t]{2}{*}{Banking} & ID & \std{84.7}{0.7} & \std{85.3}{0.0} & \std{85.0}{0.3} \\
 & OOD & \std{85.4}{1.5} & \std{86.0}{1.1} & \std{85.7}{1.2} \\
\multirow[t]{2}{*}{Finance} & ID & \std{84.2}{0.3} & \std{85.3}{0.4} & \std{84.8}{0.4} \\
 & OOD & \std{87.2}{0.8} & \std{88.5}{0.5} & \std{87.8}{0.2} \\
\multirow[t]{2}{*}{Health} & ID & \std{86.1}{0.8} & \std{87.4}{0.6} & \std{86.8}{0.1} \\
 & OOD & \std{80.4}{1.1} & \std{78.3}{0.9} & \std{79.3}{0.3} \\
\multirow[t]{2}{*}{Insurance} & ID & \std{84.8}{0.9} & \std{86.3}{0.6} & \std{85.6}{0.2} \\
 & OOD & \std{87.6}{0.2} & \std{79.4}{0.6} & \std{83.3}{0.3} \\
\multirow[t]{2}{*}{Travel} & ID & \std{85.9}{0.7} & \std{85.1}{0.4} & \std{85.5}{0.4} \\
 & OOD & \std{84.9}{1.0} & \std{80.0}{0.8} & \std{82.4}{0.3} \\
 \midrule
\multirow[c]{2}{*}{Average} & ID & \std{85.2}{0.7} & \std{85.9}{0.4} & \std{85.5}{0.3} \\
 & OOD & \std{85.1}{0.9} & \std{82.4}{0.8} & \std{83.7}{0.5} \\
\bottomrule
\end{tabular}
\caption{Dialogue act classifier cross-domain evaluation on \ours{}. In-domain (ID) training data consists of data from the same domain as the test dataset, whereas out-of-domain (OOD) training data consists of training data from the remaining (non-test) domains.}
\label{Tab:da_classifier_cv}
\end{table*}

For models trained on \ours{} to be useful in practice, they must be able to generalize beyond the limited number of domains present in \ours{}.
To measure cross-domain generalization, we perform cross validation, training separate models in which we hold out a single domain (e.g. train on Banking, evaluate on Travel).
As shown in Table~\ref{Tab:da_classifier_cv}, performance on the heldout domains is lower, particularly for recall, but does not drop substantially.

\subsection{Collection Methodology}
\label{sec:Appendix:CollectionMethodology}
\begin{table*}[tbh]
\addtolength{\tabcolsep}{-3pt}
\caption{Sample company profile for \ours (Insurance)}
\label{Tab:companyprofile}
\begin{tabular}{p{5cm}p{10cm}}
\toprule
Company name & Rivertown Insurance \\
\midrule

Domain & Insurance \\
Description & An ordinary insurance company. Provides insurance for Pets, Rent, Automobile, Life, etc. \\
Website & www.rivertowninsurance.com \\
Insurance offered & Life \\
 & Pet \\
 & Automobile \\
 & Condo \\
 & Homeowner \\
 & Renter \\
Plan Types offered (Automobile) & Basic Auto (\$1000/year) \\
 & Preferred Auto (\$1500/year) \\
 & Complete Auto (\$2000/year) \\
Plan Types offered (Condo) & Basic (\$500/year) \\
 & Condo Preferred (\$600/year) \\
Plan Types offered (Homeowner) & Basic Home (\$1200/year) \\
 & Home Preferred (\$1600/year) \\
 & Home Complete (\$2000/year) \\
Plan Types offered (Life) & Term Life Insurance (\$300/year) \\
 & Whole Life Insurance (\$1800/year) \\
 & Universal Life Insurance (\$1200/year) \\
Plan Types offered (Pet) & Petcare Basic (\$500/year) \\
 & Petcare Preferred (\$1000/year) \\
Plan Types offered (Renters) & Renters Basic (\$200/year) \\
 & Renters Preferred (\$300/year) \\
Security questions: & What is your mother's maiden name? \\
 & What is the name of your childhood best friend? \\
 & What is the name of your high school? \\
 & What is the name of your first pet? \\
 & What is the name of your favorite teacher? \\
Company's protocol to verify identity & To verify a customer's identity, you will need either: \\
& 1) FirstName, LastName, DateOfBirth, and CustomerNumber (8 digits long), or \\ 
& 2) FirstName, LastName, DateOfBirth, PhoneNumber, SocialSecurityNumber, and answer to one security question \\
\bottomrule
\end{tabular}
\end{table*}

\begin{table*}[tbh]
\addtolength{\tabcolsep}{-3pt}
\begin{tabular}{p{0.25\linewidth}p{0.4\linewidth}p{0.35\linewidth}}
\toprule
Slots & Agent suggested talking points & Customer suggested talking points \\
\midrule
& 1. Confirm the identity of the customer &  1. Ask if you can change the password on the website instead of making a phone call\\
& 2. Ask if there's anything else you can do for the customer before completing the conversation. & \\
Slots required for identity verification & To verify a customer's identity, you will need either: & \\
& 1) FirstName, LastName, DateOfBirth, and CustomerNumber (8 digits long), or & \\
& 2) FirstName, LastName, DateOfBirth, PhoneNumber,  SocialSecurityNumber, and answer to security question &  \\
SecurityQuestionAnswer &  & \\
EmailAddress & 1. Verify the email address on file. Note that the reset link will be sent to this email &  \\
\bottomrule
\end{tabular}
\caption{Sample intent/slot schema with instructions for Insurance, ResetPlan intent}
\label{Tab:sampleschema}
\end{table*}
\begin{table*}[tbh]
\addtolength{\tabcolsep}{-3pt}
\caption{Discourse complexities with examples}
\label{Tab:discourseexamples}
\begin{tabular}{lp{5.75cm}p{5.75cm}}
\toprule
Name & Description & Example \\
\midrule
ChitChat & Small talk unrelated to the intent & Hey, how's your day going? \\
Frustration & Expression of frustration & That's such a scam because you were able to fulfill this before \\
BackgroundDetail & Related information that is unnecessary for resolving the request is provided for context & \textbf{We booked a room from you last year} and were hoping to come stay again over Memorial day weekend \\
ImplicitDescriptiveIntent & A description of the problem is provided, rather than an explicit request & I got this notice in the mail regarding a rate increase but an agent told me that the rate was fixed till the end of the year \\
SlotChange & The customer amends or corrects information provided & Oh wait, my zipcode is actually 20512 now \\
IntentChange & The customer changes their request midway through a conversation. & Actually, I'd better check my balance first \\
FollowUpQuestion & The customer asks follow-up questions that are related to their original intent (can occur before the original intent has been fulfilled) & How long does it take to deliver? \\
MultiElicit & The agent elicits multiple related pieces of information at once & Could you please provide your name and date of birth? \\
MultiIntent & The customer has multiple requests & I'd also like to transfer some money while I'm here \\
MultiIntentUpfront & The customer states multiple requests at the beginning of the conversation & I wanted to check the status of a claim and add my wife to my policy \\
MultiValue & The customer provides multiple slot values, even if only a single value was requested & A: What item are you returning today? / C: I wanted to return a blender and a pan \\
Callback & The conversation is a continuation of another conversation with prior context & Hey I'm back now. I was able to restart \\
MissingInfoWorkaround & The agent requests other information because the customer does not have some requested information & C: I don't have the barcode number / A: No worries, could you give me your last name? \\
MissingInfoCantFulfill & The agent is uable to fulfill a request because the customer is missing some information & C: I don't have the barcode number / A: I will need that number to fulfill this request for you, so you'll need to call back once you find it \\
PauseForLookup & Customer or agent asks the other party to hold or wait while they look up some information. & Can you hold on while I find that number? \\
SlotLookup & The agent already has information about the customer and only needs to verify details & Is johndoe@gmail.com still a good email address for you? \\
Overfill & The customer provides more information than asked & A: What's your first name? / C: It's John Doe, email is johndoe@gmail.com \\
SlotCorrection & The agent corrects a customer regarding a product or service offered by the company & C: I ordered the Derek pot set / A: The Darin set, yes I see your order here. \\
\bottomrule
\end{tabular}
\end{table*}

\begin{table*}[tbh]
\addtolength{\tabcolsep}{-3pt}
\caption{Spoken complexities with examples}
\label{Tab:spokenexamples}
\begin{tabular}{lp{5.75cm}p{5.75cm}}
\toprule
Name & Description & Example \\
\midrule
Backchanneling & Speaker(s) interrupts to signify that they are listening & uh huh, right go on \\
Confirmation & Speaker(s) confirms their understanding of what the other party says, either due to being unable to hear properly or as an acknowledgement & you said thirty dollars for everything? \\
Disfluencies & Speaker(s) exhibit disfluencies and unfinished thoughts & No no that it isn't it. You'll need to get the um first can you see in the right hand corner your username? \\
HesitationsFillersHigh & Speaker(s) includes a high percentage of filler words to mark a pause or hesitation & um, uh, er \\
HesitationsFillersLow & Speaker(s) includes a low percentage of filler words to mark a pause or hesitation & um, uh, er \\
Interruption & Speaker(s) interrupts, especially when it is unclear when a thought has ended & C: Yeah I wanted to get help with filling out these forms and then um / A: Okay sure I can help with that / C: I'm also going to want to change some personal information \\
PartialInformation & Speaker(s) is only able to give partial information due to forgetfulness & I'm trying to get a hold of that thing, you know the 228 something form \\
Rambling & Speaker(s) ramble and repeat themselves. They may paraphrase themselves & Oh I got it I see. So basically I had this employee who just up and moved and it's been a disaster so I don't know um like where to even find him, but he was in charge of all the records so things are just a huge mess and I don't know how to update this it's been impossible and I don't really understand everything that goes into this. \\
AskToRepeatOrClarify & One party is unable to hear and either expresses this or asks for repetition or clarification. & Did you say that's 128 or 1228? \\
Spelling & Speakers should spell things out if they need to clarify uncommon or ambiguous spellings & That's Jon spelled without an H. \\
\bottomrule
\end{tabular}
\end{table*}

Table~\ref{Tab:companyprofile} provides an example company profile used to assist in achieving cross-dataset consistency.
Table~\ref{Tab:sampleschema} provides an example intent/slot schema for a ResetPassword intent in the Insurance domain.
Table~\ref{Tab:discourseexamples} provides example utterances for each discourse complexity.
Table~\ref{Tab:spokenexamples} provides examples of spoken complexities used to simulate spoken-form conversations in the \oursinsuranceself{}.

\begin{figure*}[tbh]
\centering
\begin{framed}
Spelling (80\%),
HesitationsFillersLow (60\%),
Backchanneling (50\%),
HesitationsFillersHigh (40\%),
Interruption (40\%),
Confirmation (30\%),
Disfluencies (30\%),
Rambling (30\%),
AskToRepeatOrClarify (25\%),
PartialInformation (10\%)
\end{framed}
\caption{Complexities present in spoken form conversations and corresponding target percentages, estimated based on manual inspection of a small set of real conversations.}
\label{Fig:spoken}
\end{figure*}

Figure~\ref{Fig:spoken} provides target percentages for complexities used to simulate spoken conversations in \oursself{}.

\subsection{Conversation Examples}
\begin{table*}[tbh]
\centering
\addtolength{\tabcolsep}{-3pt}
\scalebox{0.90}{
\begin{tabular}
{m{0.5\linewidth}>{\centering\arraybackslash}m{0.25\linewidth}>{\centering\arraybackslash}m{0.25\linewidth}}
\toprule
 Text & Annotations & Complexities \\
\midrule
 \textbf{A}: Good afternoon. Welcome to Intellibank. This is Rose. How can I help you today? & \ann{ElicitIntent} \\
 \textbf{C}: Hi Rose, my name is \labelbox{Kim}{FirstName}. I have a couple of questions for you actually I need to make a wire transfer and I also have an issue with my card. I may have lost it. So, but I'd like to to do the wire transfer first and then well maybe we can re-issue me a new card. & \ann{InformSlot}, \ann{InformIntent} \newline (\ann{ExternalWireTransfer}, \ann{ReportLostStolenCard}, \ann{RequestNewCard}) & MultiIntentUpfront, Overfill \\
 \textbf{A}: OK. & & Backchanneling \\
 \textbf{A}: OK, certainly. I can help you with that issue. Kim, first I need to know your \labelbox{last name}{LastName}. & \ann{ElicitSlot} \\
 \textbf{C}: Sure, it's \labelbox{Johns}{LastName}, \labelbox{J O H N S}{LastName}. & \ann{InformSlot} & Spelling \\
 \textbf{A}: All right, thank you so much Ms. Johns. And also, what is your \labelbox{date of birth}{DateOfBirth}, please? & \ann{ElicitSlot} \\
 \textbf{C}: It's \labelbox{April twenty-fourth nineteen seven seven}{DateOfBirth}. & \ann{InformSlot} \\
 \textbf{A}: OK, thank you so much. and I also need the \labelbox{account number}{AccountNumber} that you would like to make the transfer from. & \ann{ElicitSlot} \\
 \textbf{C}: sure, hold on for one moment. Let me go grap that. & & PauseForLookup \\
 \textbf{A}: Sure, take your time. & \\
 \textbf{C}: All right, I have it right here now. & \\
  \multicolumn{3}{c}{...}  \\
\bottomrule
\end{tabular}
}
\caption{Example conversation from \ours{} \oursbanking{}. The initial customer utterance is labeled with multiple intents (ExternalWireTransfer, ReportLostStolenCard, and RequestNewCard).}
\label{tab:example_conversation_spoken_1}
\end{table*}
\begin{table*}[tbh]
\centering
\addtolength{\tabcolsep}{-3pt}
\scalebox{0.90}{
\begin{tabular}
{m{0.5\linewidth}>{\centering\arraybackslash}m{0.25\linewidth}>{\centering\arraybackslash}m{0.25\linewidth}}
\toprule
 Text & Annotations & Complexities \\
\midrule
  \textbf{A}: Thank you for calling Intellibank. My name is Izumi. \labelbox{Who}{Unspecified} am I speaking with today? & \ann{ElicitSlot} \\
 \textbf{C}: Hey, Izumi. My names \labelbox{Edward}{FirstName} \labelbox{Elric}{LastName}. just calling in cuz I have a big problem today. I was at the I was at a couple of stores with my wife buying supplies for a house and I think I lost my card. I can't remember when I lost it since we were at one of the stores. My wife actually used one of her cards to get the store points that they offer. don't know if it was before then or after. I think it was after because we went to go eat somewhere there. just calling in to get some help. & \ann{InformIntent}, \ann{InformSlot} \ann{(ReportLostStolenCard)} & BackgroundDetail, ImplicitDescriptiveIntent \\
 \textbf{A}: Oh, no. I'm so sorry to hear that Mr. Elric. I know that it's difficult when you lose your your bank card. but yeah. Please rest assured, I will do everything I can to make sure that your account is secure and then we'll get you a new card as well as soon as possible, OK? & & ChitChat \\
 \textbf{C}: Oh, sweet. Never lost my card before. Kind of worried. So not sure what I have to do. Is there anything that I need to give you first so I can get my card? & \ann{InformIntent} & Frustration, FollowUpQuestion \\
 \textbf{A}: Yes, Mr. Elric. So first of all, I'll just need to ask for your personal information so that I can pull up your account . can I get your \labelbox{account number}{AccountNumber}, please? & \ann{ElicitSlot} \\
 \textbf{C}: Yeah. It is \labelbox{one zero zero, one seven zero}{AccountNumber} and the last four of my social are \labelbox{fifty-two fourteen}{LastFourSocial} in case you need that. & \ann{InformSlot} & Overfill \\
 \textbf{A}: yes. Thank you for that information. And then, I just wanna make sure that I have that account number correct. So I have \labelbox{one zero zero, one seven zero}{AccountNumber}. And then, the last four of the social were \labelbox{five two one four}{LastFourSocial}. & \ann{ConfirmSlot} & \\
 \textbf{C}: Yeah yeah. That's right. & & \\
 \textbf{A}: Perfect. And then, can I also verify your \labelbox{date of birth}{DateOfBirth}, please? & \ann{ElicitSlot} \\
 \textbf{C}: Yeah. It is \labelbox{October fourth, nineteen ninety}{DateOfBirth}. & \ann{InformSlot} \\
 \textbf{A}: OK, perfect. Thank you so much. Let me just pull up your account. Give me one moment. & & PauseForLookup \\
 \textbf{C}: Oh, OK. & & \\
 \textbf{A}: All right. Just one second. & & PauseForLookup  \\
 \textbf{C}: Yeah. I really hope get help to find my card. We are renovating our house at the moment right now. Started redoing our walls not too long ago. It's a bunch of wallpaper, so we just need help finishing removing it. And then, my wife is gonna head off to the store to get some paint to start that project. & & BackgroundDetail, ChitChat\\
 \textbf{A}: Oh, that's awesome. Do you guys have like a color scheme or color palette that you're working with? & & ChitChat \\
  \multicolumn{3}{c}{...}  \\
\bottomrule
\end{tabular}
}
\caption{Example conversation from \ours{} \oursbanking{} with discourse complexities (e.g. BackgroundDetail, ChitChat, and FollowUpQuestion). In this conversation, the customer provides considerable background information related, but not necessary for understanding their intent.}
\label{tab:example_conversation_spoken_2}
\end{table*}
\begin{table*}[tbh]
\centering
\addtolength{\tabcolsep}{-3pt}
\scalebox{0.90}{
\begin{tabular}
{m{0.5\linewidth}>{\centering\arraybackslash}m{0.25\linewidth}>{\centering\arraybackslash}m{0.25\linewidth}}
\toprule
 Text & Annotations & Complexities \\
\midrule
 \textbf{A}: Hello and thank you for calling Intellibank. This is Mark speaking. How could I help you today? & \ann{ElicitIntent} \\
 \textbf{C}: Hi Mark. My name is \labelbox{Dorothy}{FirstName} \labelbox{Lee}{LastName}. I would like to check my \labelbox{savings}{TypeOfAccount} account balance. & \ann{InformSlot}, \ann{InformIntent} (\ann{CheckAccountBalance}) & Overfill\\
 \textbf{A}: OK Dorothy. I can help you with that if you give me one second. Could you in the meantime give me your \labelbox{date of birth}{DateOfBirth} please? & \ann{ElicitSlot} \\
 \textbf{C}: Yeah. It's \labelbox{April the fifteenth nineteen ninety-nine}{DateOfBirth}. & \ann{InformSlot} \\
 \textbf{A}: OK perfect. Thank you so much. Now also could you give me your \labelbox{account number}{AccountNumber}? & \ann{ElicitSlot} \\
 \textbf{C}: Oh no I don't have it with me. Is that a problem? & \\
 \textbf{A}: That shouldn't be a problem. do you happen to have your \labelbox{credit card number}{CreditCardNumber} on you as well? & \ann{ElicitSlot} & MissingInfoWorkaround \\
 \textbf{C}: I don't have that either. & \\
 \textbf{A}: OK well I need some form of identifying you. do you happen to have a \labelbox{driver's license}{OtherIDNumber} or another state ID issued number? & \ann{ConfirmSlot}, \ann{ElicitSlot} & MissingInfoWorkaround \\
 \textbf{C}: No sorry I I didn't bring anything with me today. Is there any other way we can do it? & \ann{InformIntent} &  Disfluencies \\
\textbf{A}: unfortunately I'm gonna need some of that information to to process your request. so unfortunately because there's a lot of of theft going on I I it's could be fraud. I'm not sure that you are who you say who you are and if you can't give me that information. We use those as security checkpoints then I won't be able to complete your request for you. I apologize for that Mrs. Dorothy Lee. & & MissingInfoCantFulfill, Disfluencies \\
 \textbf{C}: Well I'm kind of disappointed because I always get like a terrible service customer here but OK. I want you to help me with another thing. Is that possible? & & Frustration, MultiIntent \\
 \textbf{A}: Yes of course. what what else could I help you with today? . & \ann{ElicitIntent} \\
  \multicolumn{3}{c}{...}  \\
\bottomrule
\end{tabular}
}
\caption{Example conversation from \ours{} \oursbanking{} with discourse complexities (e.g. MultiIntent and MissingInfoWorkAround). In this conversation, the customer is unable to provide the necessary information for identity verification, despite the agent offering multiple possible workarounds.}
\label{tab:example_conversation_spoken_3}
\end{table*}

\begin{table*}[tbh]
\centering
\addtolength{\tabcolsep}{-3pt}
\scalebox{0.90}{
\begin{tabular}
{m{0.5\linewidth}>{\centering\arraybackslash}m{0.25\linewidth}>{\centering\arraybackslash}m{0.25\linewidth}}
\toprule
 Text & Annotations & Complexities \\
\midrule
 \textbf{A}: Thank you for calling Rivertown Insurance. How may I help you today? & \ann{ElicitIntent} \\
\textbf{C}: Yes. I need to do something about lowering my premiumiss I recently lost my job and just can't afford the payments anymore. I don't want to change companies. I've been with you guys for years. & \ann{InformIntent} (\ann{ChangePlan}, \ann{RequestDiscount}) & BackgroundDetail \\
 \textbf{A}: I'm sorry to hear you lost your job. Let's see what we can do. & & \\
 \textbf{C}: Okay. Thank you. & \\
 \textbf{A}: May I have your \labelbox{first}{FirstName} and \labelbox{last name}{LastName}? & \ann{ElicitSlot} & MultiElicit \\
 \textbf{C}: It's \labelbox{Maria}{FirstName} \labelbox{Sanchez}{LastName}. & \ann{InformSlot} \\
 \textbf{A}: Thank you, Maria. Do you happen to have your \labelbox{customer number}{CustomerID}? & \ann{ConfirmSlot}, \ann{ElicitSlot} \\
 \textbf{C}: I think so. Let me check my purse. & & PauseForLookup \\
 \textbf{A}: Okay. Take your time. & \\
 \textbf{C}: It's \labelbox{one two three four five six seven eight}{CustomerID}. & \ann{InformSlot} \\
 \textbf{A}: Perfect, and can you verify your \labelbox{date of birth}{DateOfBirth} please? & \ann{ElicitSlot} \\
 \textbf{C}: It's \labelbox{seven twenty six nineteen eighty nine}{DateOfBirth}. & \ann{InformSlot} \\
 \textbf{A}: Thank you, Maria. I have you pulled up here. \labelbox{Which}{PlanType} policy were you looking at reducing the payment on? Life or Auto? & \ann{ConfirmSlot}, \ann{ElicitSlot} & SlotLookup \\
 \textbf{C}: the \labelbox{auto policy}{PlanType}. I don't want to change my life insurance. & \ann{InformSlot} \\
 \textbf{A}: Okay. It looks like you're on the \labelbox{Complete plan}{PlanName}. Does that sound correct? & \ann{ConfirmSlot}, \ann{ElicitSlot} & SlotLookup \\
 \textbf{C}: Yes it was the highest one. & \\
 \textbf{A}: Okay. We do have two options with lower payments how much lower did you need to go? & \ann{ElicitSlot} \\
  \multicolumn{3}{c}{...}  \\
\bottomrule
\end{tabular}
}
\caption{Example conversation from \oursself{} demonstrating primarily discourse complexities.}
\label{tab:example_conversation_written_1}
\end{table*}
\begin{table*}[tbh]
\centering
\addtolength{\tabcolsep}{-3pt}
\scalebox{0.90}{
\begin{tabular}
{m{0.5\linewidth}>{\centering\arraybackslash}m{0.25\linewidth}>{\centering\arraybackslash}m{0.25\linewidth}}
\toprule
 Text & Annotations & Complexities \\
\midrule
 \textbf{A}: Hello, Rivertown Insurance, Marissa here how could I help? & \ann{ElicitIntent} \\
 \textbf{C}: Yep I just made my account and now it says call to enroll? & \ann{InformIntent} \ann{(EnrollInPlan)} & Callback, BackgroundDetail \\
 \textbf{A}: Mhm I could enroll you in one of our plans over the phone. & & \\
 \textbf{C}: Uh-huh. & & Backchanneling \\
 \textbf{A}: Do you have any idea what sort of \labelbox{plan}{PlanType} you wanted to enroll i-in? & \ann{ElicitSlot} & Disfluencies \\
 \textbf{C}: Nah could I also get a quote over the phone? & \ann{InformIntent} \ann{(GetAutoQuote)} & MultiIntent \\
 \textbf{A}: Mhm I could also do that for you sir. & &  \\
 \textbf{C}: Oh great Marissa thank you. & \\
 \textbf{A}: Mhm no problem sir, could I get your \labelbox{first and last name}{FirstName}\labelbox{first and last name}{LastName} p-please? & \ann{ElicitSlot} & Disfluencies \\
 \textbf{C}: \labelbox{Jakob}{FirstName} \labelbox{Burbert}{LastName}. & \ann{InformSlot} \\
 \textbf{A}: Jacob \labelbox{j.a.c.o.b}{FirstName}? & \ann{ConfirmSlot} & Spelling, Confirmation \\
 \textbf{C}: Nah \labelbox{Jakob with a k}{FirstName}. & \ann{InformSlot} & Spelling \\
 \textbf{A}: Got it is Burbert \labelbox{b.u.r.b.e.r.t}{LastName}? & \ann{ConfirmSlot} & Spelling, Confirmation \\
 \textbf{C}: Yep! & \\
 \textbf{A}: Oh great so I'm guessing you don't have your customer number? & \ann{ElicitSlot} \\
 \textbf{C}: Nah I don't have that what do I need it for? & \ann{InformIntent} \\
 \textbf{A}: Huh it's fine I can just use some other infomation to complete the ID check. & & HesitationsFillersLow, MissingInfoWorkaround \\
 \textbf{C}: Whew alright. & \\
 \textbf{A}: Could I get your \labelbox{date of birth}{DateOfBirth}, \labelbox{phone number}{PhoneNumber} and \labelbox{social security}{SocialSecurityNumber}? &  \ann{ElicitSlot} & MultiElicit \\
 \multicolumn{3}{c}{...}  \\
\bottomrule
\end{tabular}
}
\caption{Example conversation excerpt from \oursself{} demonstrating both spoken complexities as well as discourse complexities.}
\label{tab:example_conversation_written_2}
\end{table*}
\begin{table*}[tbh]
\centering
\addtolength{\tabcolsep}{-3pt}
\scalebox{0.90}{
\begin{tabular}
{m{0.5\linewidth}>{\centering\arraybackslash}m{0.25\linewidth}>{\centering\arraybackslash}m{0.25\linewidth}}
\toprule
 Text & Annotations & Complexities \\
\midrule
 \textbf{A}: Rivertown Insurance, this is Carla speaking how can I help you? & \ann{ElicitIntent} \\
 \textbf{C}: Um hi could I create an account and enroll in a plan over the phone? & \ann{InformIntent} (\ann{CreateAccount}, \ann{EnrollInPlan}) & HesitationsFillersHigh, MultiIntentUpfront \\
 \textbf{A}: Mhm yes sir I can help you with that. Do you want to start the sign up now? & \ann{ElicitIntent} \\
 \textbf{C}: Uh-huh can I only sign up over the phone? & \ann{InformIntent} \ann{(AskFAQ)} & FollowUpQuestion \\
 \textbf{A}: Oh, no you could also use our website to create your account if you & \\
 \textbf{C}: Eee I think I'll just have you help me . & \ann{InformIntent} & HesitationsFillersHigh, Interruption \\
 \textbf{A}: Sure thing sir first can I get your \labelbox{first and last name}{FirstName}\labelbox{last name}{LastName} please? & \ann{ElicitSlot} & MultiElicit \\
 \textbf{C}: Uh yeah \labelbox{Jonny}{FirstName}, j.o.n.n.y. & \ann{InformSlot} & Spelling \\
 \textbf{A}: Mhm. & & Backchanneling \\
 \textbf{C}: \labelbox{Barbados}{LastName} \labelbox{b.a.r.b.a.d.o.s.}{LastName} & \ann{InformSlot} & Spelling \\
 \textbf{A}: Uh-huh thank you could I get your \labelbox{phone number}{PhoneNumber} now? & \ann{ElicitSlot} \\
 \textbf{C}: Mhm \labelbox{five four three}{PhoneNumber}. & \ann{InformSlot} \\
 \textbf{A}: \labelbox{Five four three}{PhoneNumber}. & \ann{ConfirmSlot} & Confirmation \\
 \multicolumn{3}{c}{...}  \\
\bottomrule
\end{tabular}
}
\caption{Example annotated conversation excerpt from \oursself{} demonstrating both spoken complexities (e.g. Backchanneling) as well as discourse complexities (e.g. MultiIntentUpfront and FollowUpQuestion).}
\label{tab:example_conversation_written_3}
\end{table*}

Tables~\ref{tab:example_conversation_spoken_1}, \ref{tab:example_conversation_spoken_2}, and \ref{tab:example_conversation_spoken_3} provide examples of conversations collected in \oursspoken{}.
Tables~\ref{tab:example_conversation_written_1}, \ref{tab:example_conversation_written_2}, and \ref{tab:example_conversation_written_3} provide examples of conversations collected in \oursself{}.

\end{document}